\documentclass[review,5p, times,twocolumn]{elsarticle}

\usepackage{url}  
\usepackage[dvipsnames]{xcolor}
\usepackage{tabularx}

\usepackage{amssymb}

\makeatletter
\def\ps@pprintTitle{%
  \let\@oddhead\@empty
  \let\@evenhead\@empty
  \def\@oddfoot{\reset@font\hfil\thepage\hfil}
  \let\@evenfoot\@oddfoot
}
\makeatother

\journal{Online Social Networks and Media}

\begin{document}

\begin{frontmatter}

\title{Empowering NGOs in Countering Online Hate Messages}

\author[label1,label2]{Yi-Ling Chung}
\ead{ychung@fbk.eu}
\author[label1]{Serra Sinem Tekiro\u{g}lu}
\ead{tekiroglu@fbk.eu}
\author[label1]{Sara Tonelli}
\ead{satonelli@fbk.eu}
\author[label1]{Marco Guerini}
\ead{guerini@fbk.eu}
\address[label1]{Fondazione Bruno Kessler, Via Sommarive 18, Povo, Trento, Italy}
\address[label2]{University of Trento, Italy}

\author{}

\address{}

\begin{abstract}
Studies on online hate speech have mostly focused on the automated detection of harmful messages. Little attention has been devoted so far to the development of effective strategies to fight hate speech, in particular through the creation of counter-messages. 
While existing manual scrutiny and intervention strategies are time-consuming and not scalable, advances in natural language processing have the potential to provide a systematic approach to hatred management. In this paper, we introduce a novel ICT platform that NGO operators can use to monitor and analyze social media data, along with a counter-narrative suggestion tool.
Our platform aims at increasing the efficiency and effectiveness of operators' activities against islamophobia.
We test the platform with more than one hundred NGO operators 
in three countries through qualitative and quantitative evaluation. Results show that NGOs favor the platform solution with the suggestion tool, and that the time required to produce counter-narratives significantly decreases.
\end{abstract}


\end{frontmatter}

\section{Introduction}
\label{intro}

{T}{he} rapid growth of social media platforms makes communication and interaction easier and faster. Still these platforms can also become a fertile ground for hatred content \cite{Mondal:2017:MSH:3078714.3078723}. While a huge number of users passively receive and digest news and messages from social media, some can actively generate inflaming comments and post them publicly on any subject while remaining anonymous \cite{aguilera2016islamonausea}. The side effect is that hate groups may use social media platforms as a propaganda tool to spread hate speech, appeal to a wider audience, and marginalize specific communities \cite{awan2016islamophobia, hall2013hate}. 
This scenario has motivated authorities of many countries in the search for effective ways of fighting online hatred. For example, UNESCO published a report on countering online hatred \cite{gagliardone2015countering}, and the Council of Europe coordinated the No Hate Speech Movement\footnote{http://www.nohatespeechmovement.org/} and Get The Trolls Out project\footnote{https://www.getthetrollsout.org} to address hate speech.

Countering online hate involves two main challenges: monitoring and intervention. The first task involves identifying hateful messages, collecting information on who posts them and try to understand topics, trends and communities of haters from a large amount of social media data. The second includes several possible actions to concretely counter online hate speech. Both activities are challenging, since it is extremely difficult during monitoring to identify potential online threats in real-time and to further stop them from spreading. Standard approaches mainly rely on users' reporting activity that can lead to the deletion of hate content or suspension of haters' accounts. Still, manual intervention cannot govern the magnitude of the problem, and empowering censorship on social media may limit the right to freedom of speech and diverse opinions. 
An alternative and novel strategy is to directly intervene in the discussion with counter-narratives. Counter-narratives (CNs) are non-negative responses to hate speech, targeting and counteracting extreme statements through fact-bound arguments or alternative perspectives \cite{benesch2014countering, schieb2016governing}. This approach is preferable for several reasons: the right to freedom of speech is preserved, stereotypes and misleading information are rectified with reliable evidence, the exchange of diverse viewpoints is encouraged (also among possible bystanders) \cite{benesch2014countering, schieb2016governing}. 

Not all responses to a hate message can be seen as `good' counter-narratives. Following NGO’s guideline (e.g., the Get The Trolls Out project), a good counter narrative is a response that provides non-negative feedback with fact-bound arguments. Prior work on Twitter \cite{benesch2016}  shows that effective counter narratives, inducing favorable impact on hate users, often use strategies including warning of consequences, empathy, affiliation, and humor. We show an example below, where a hate-message (HM) is reported with two possible responses. The first, R1, is a non-negative and appropriate response, which we consider in this work as a good counter-narrative. The second response, instead, is not appropriate because it contains abusive language, and it therefore does not represent the kind of counter-speech that should be used to address online hatred:

        {\textbf{HM: }}{\textit{The world would be a better place without Muslims. They are only killing and raping our children.}}\label{HM}
        
        {\textbf{R1: }}{\textit{The world would actually be a very dark place without Muslims who contribute a lot to our society. What about our Muslim doctors, scientists, philanthropists, job-creators?} }\label{CN}

        {\textbf{R2: }}{\textit{You are very stupid and dumb to speak negatively about Muslims. Hell is where you belong!} }\label{CN}

Currently, several NGOs are engaging in hate countering activity on social media platforms. They typically rely on volunteers and operators who, after a specific training, learn to (i) detect harmful conversations, (ii) identify relevant hate accounts and (iii) either report the hate content or produce effective counter-narratives, in most cases using fake profiles for safety reasons. This manual intervention strategy is clearly not feasible on a large scale, since it would become too costly and time-consuming for NGOs to handle. We therefore propose in this work a novel online platform that enhances NGO operators' online activities (similar to the concept of `digital augmentation' presented in \cite{stella2019influence}), by automatizing many tasks in monitoring and counter-narratives writing. This is done through a combination of natural language processing, social network analysis and advanced data visualisation tools.

While several research works have addressed the automatic detection of online hate speech using natural language processing \cite{basile-etal-2019-semeval,DBLP:conf/konvens/StrussSRWK19,DBLP:conf/semeval/ZampieriNRAKMDP20,DBLP:conf/evalita/SanguinettiCNFS20}, limited studies have explored automation of hatred countering \cite{qian2019benchmark, tekiroglu2020generating} due to the scarcity of hate-speech / counter-narrative pairs. The goal of employing a suggestion tool lies in providing automated responses that reduce the time and efforts of NGO operators in responding to offensive posts, thus yielding more systematic and sustainable impact against online hate speech. Furthermore, a partial automation of the `seek and reply' task of NGO operators would alleviate the emotional burden of having to interact with hateful content for hours, a situation that has proven to affect with lasting psychological and emotional harm content moderators exposed to extreme content \cite{riedl2020downsides}.  The issue of moderators' well-being has been tackled in the past via automatic image modification, for example image blurring \cite{dang2018but} or grayscaling \cite{karunakaran2019testing}. To our knowledge, no previous work has addressed the problem by automating the creation of hate countering messages through an interactive and real-time interface.

The novel ICT platform presented in this work has been designed to systematically support NGOs in monitoring, analysing and countering online hatred in English, French, and Italian, focusing on Islamophobic messages.
The platform includes two main features: the first allows users to promptly monitor hate speech on Twitter targeting Muslims and Islam. Then, users can rely on the second feature, a novel counter-narrative suggestion tool, to automatically compose suitable responses to hate content. The platform was tested by 112 operators across 3 different countries through an extensive evaluation that includes open-ended questions, a user experience questionnaire, and time measurement. The experimentation shows that the platform can effectively empower operators in the crucial task of hate countering. 

\textit{NOTE}:  This  paper  contains  examples  of  language  which  may  be  offensive  to  some  readers. They do not represent the views of the authors.

\section{Background and Related work}
In this section we briefly review the main works related to hate speech in terms of monitoring and countering.

\subsection{Online Hate Speech Monitoring}
Online hate speech monitoring is a crucial task that is difficult to carry out manually due to the severe magnitude and scale of the problem. We identify two types of monitoring: real-time and retrospective. The former aims to provide continuous monitoring of social media, and it is more difficult to achieve because of the ever-changing nature of social media, communities and shared contents. Among the best known projects dealing with real-time monitoring there is the Umati Project \cite{eisf2014}, concerning the run-up to Kenya's general elections in March 2013. It involved several operators that manually checked selected blogs, forums, online newspapers, Facebook, and Twitter daily, in English and seven other languages, to track and report hate speech. Also the COSMOS platform \cite{burnap2015cosmos}  offers real-time data collection from Twitter and data analysis both at tweet level, such as user sentiment, gender, and geospatial location analysis, and at corpus level, including keyword frequency analysis and hashtag usage. Despite the relevance of the project, however, human error was an issue and scaling up the monitoring operations was reportedly expensive. Other similar projects have been carried out around the world, including Myanmar, Nigeria, Sri Lanka, Ethiopia, Czech Republic and the US. In some cases, toolkits and guidelines developed within the Dangerous Speech project  \cite{dangspe} have been used, with the goal to classify and analyse hate speech and to diminish its harmful impact. 
However, no specific platform to support the monitoring process has been developed within this initiative.

More recent hate monitoring platforms feature classification and visualization of hate speech applying machine learning techniques. For instance, Contro l’odio \cite{capozzi2019computational} focuses on detecting anti-immigrants discourse in Italy and providing the temporal and spatial reports regarding hate messages on Twitter. Similarly, the system introduced in \cite{menini2019system} supports the educators in high schools by identifying cyberbullying phenomena on Instagram and depicting the network of interest. MANDOLA \cite{paschalides2020mandola} is a platform that emphasizes the capability of performing large-scale classification and illustration of hate activities. Although the aforementioned tools conduct analysis on hate incidents, as opposed to the platform introduced in this work, their functionalities are served for specific languages: English or Italian.

While for real-time monitoring we could not identify any ICT platform used by institutions and NGO activists to automatically track hate speech in real scenarios, more works exist related to retrospective monitoring, probably because storing and cleaning data before analysing them make the output more accurate and visualisations can be better controlled. Monitoring platforms have been developed so that users explore hate speech datasets after they have been collected and analysed with natural language processing techniques \cite{Capozzi:2018:DVP:3240431.3240437,Mondal:2017:MSH:3078714.3078723,menini2019system,quteprints101369,doi:10.1002/poi3.85,mandola}.
To our knowledge, no platform exists that provides both real-time and retrospective monitoring like the one presented in this paper. Furthermore, no existing monitoring platforms were also designed to respond to incidents. 

\subsection{Counter-narratives against Hate Speech}
Existing strategies for combating hate speech on social media are  mostly limited to content deletion, shadow banning and account suspension. A few recent works have proposed an alternative approach, i.e. responding to hate speech using counter-narratives that users can manually create following some guidelines and best practices  \cite{bartlett2015counter, gagliardone2015countering, richards2000counterspeech}.
Several works examined the effects of counter-narratives on social media, such as Facebook \cite{schieb2016governing}, Twitter \cite{wright2017vectors, benesch2014countering}, and Youtube \cite{ernst2017hate, mathew2018thou}, and proved that they are indeed effective in fighting online hate \cite{schieb2016governing, silverman2016impact, ernst2017hate, munger2017tweetment, wright2017vectors}. 
Also, counter-narrative datasets have been created for research purposes \cite{chung2019conan, mathew2018analyzing, qian2019benchmark}. While using counter-narratives is an advised approach, composing a counter-narrative is time-consuming and demanding of wide knowledge and training. Although this training is part of the activities of NGOs who want to fight online hate, this is generally not sustainable on a large scale. A traditional scenario for an expert operator in NGOs to compose a counter-narrative includes reading hate messages, searching for and filtering related information online, and composing a counter-narrative. While a few recent studies have attempted to automatize the creation of counter messages using natural language generation techniques \cite{qian2019benchmark, tekiroglu2020generating, chung2020italian}, to the best of our knowledge, no work has tried to integrate this process in the workflow of NGO operators and evaluate it in a real scenario. In this work, we therefore aim at filling this gap by shortening the time and effort needed for locating hate content and composing proper counter-narratives.

\section{A Platform against Islamophobic Discourse}
We present in this section the online platform we developed and its two main functionalities: islamophobic speech \textit{monitoring} and \textit{countering}. NGO operators can access the platform using a password-protected account. The monitoring functionalities focus on Twitter, a choice that is mainly motivated by restrictions by other social media platforms. Indeed, neither Facebook nor Instagram, two major players dealing with online hate speech, make their APIs  available to researchers for data collection and analysis. Another reason why we focus on Twitter is that it is the social media platform that removes the least content after users report hateful messages, i.e. only 43.5\% of reported messages are deleted compared to 71.7\% by YouTube, Facebook, and Instagram on average\footnote{\url{https://ec.europa.eu/commission/news/countering-illegal-hate-speech-\online-2019-feb-04_en}}. Consequently, the issue of automatic hate speech detection on Twitter becomes particularly urgent and relevant. 

The countering functionalities, instead, can either take Twitter data as input through a link between the two functionalities, or process any message pasted or written by NGO operators. The input message calls the underlying counter-narrative engine, which suggests possible responses to the hateful post.

\subsection{Monitoring Functionalities}
The three NGOs that were involved in this study devote part of their activities to monitor islamophobia on social media platforms. In particular, they are interested in tracking ongoing discourse against Muslims, understanding which communities are more active in spreading hateful messages, and monitoring the events that trigger peaks in islamophobic messages. While designing the monitoring functionalities of the platform, we took these requirements into account, trying to display analytics that operators would not be able to obtain manually because they cross-reference and summarise information extracted from large amounts of data. On the other hand, operators need also a fine-grained view of specific days, terms or users, therefore requiring a close -- distant reading paradigm \cite{moretti-13}.

In order to account for all these aspects, the monitoring functionalities cover both real-time Twitter activity and past  messages, collected between October 2018 and May 2019. This time span was bound to the duration of the collaboration with the NGOs within the HATEMETER\footnote{\url{http://hatemeter.eu/}} project, but it could potentially be extended given that the monitoring process is fully automatic, and tweets of interest could be saved daily for future use. The analyses are available for English, French and Italian data, which were filtered by using the `language' tag provided by the Twitter API. As a preliminary step, a list of around 20 keywords and hashtags related to islamophobic discourse was collected from social scientists and NGO operators for each of the three languages. These include for example \#banislam, \#stopIslam, \#NoMoschee and \#IslamHorsDEurope. The keywords and hashtags have been used to query the Twitter APIs (\url{https://developer.twitter.com/en/docs/api-reference-index}) and retrieve content containing at least one of the hashtags/keywords of interest over a period of time starting from October 2018. A summary of the monitored data is reported in Table \ref{tab:stats}, showing the number of tweets containing at least one islamophobic term or hashtag, divided into original messages, retweets or replies for each language.

\begin{table}[h]
\caption{Islamophobic Messages Monitored between October 2018 and May 2019 in English, French and Italian}
\label{tab:stats}
\centering
\begin{tabular}{|l|r|r|r|r|}
\hline
             & \textbf{Tweets} & \textbf{Replies} & \textbf{Retweets} & \textbf{Total} \\ \hline
\textbf{En}  & 108,062         & 94,874           & 1,004,550         & 1,207,486      \\ \hline
\textbf{Fr}  & 151,738         & 268,548          & 1,206,347         & 1,626,633      \\ \hline
\textbf{Ita} & 35,569          & 27,562           & 213,713           & 286,844        \\ \hline
\end{tabular}
\end{table}

\begin{figure*}[h!]
  \centering
  \includegraphics[width=1\textwidth]{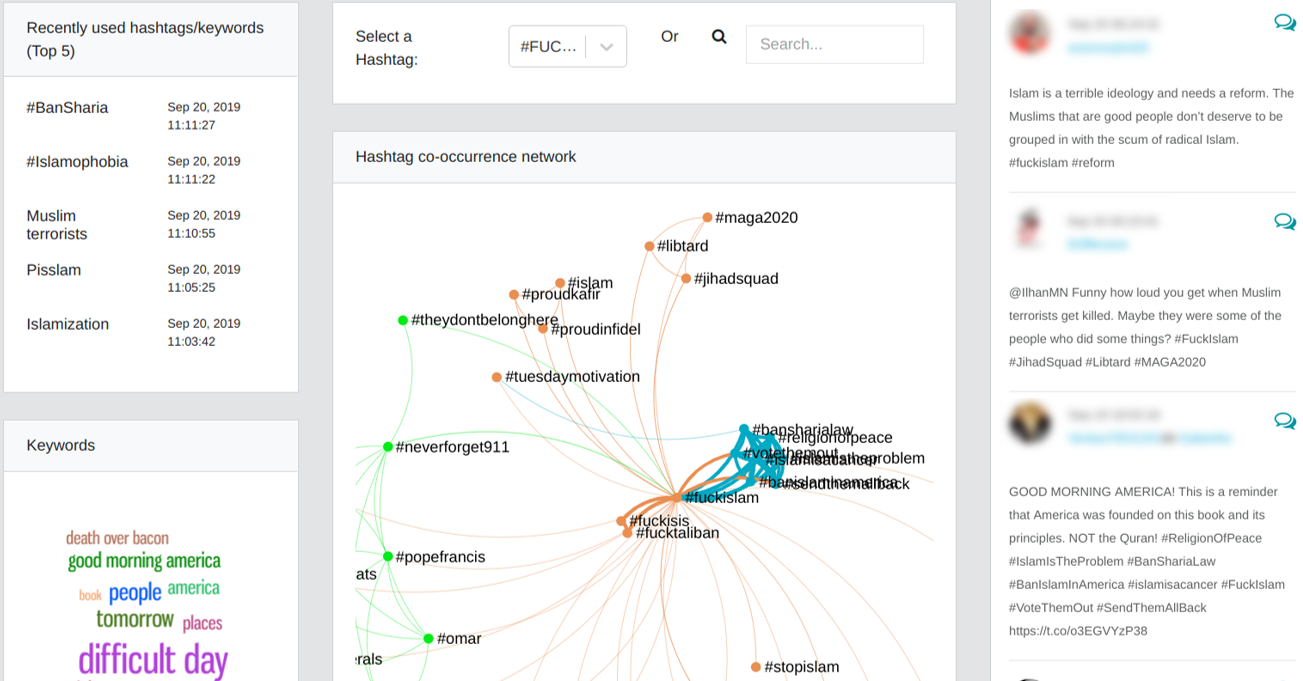}
  \caption{Screenshot of the real-time monitoring view: after selecting the \#FUCKISLAM hashtag, the network of co-occurring hashtags is displayed, together with the most recent tweets containing it. A small bubble button can be clicked to direclty obtain counter narratives (CNs)} to that particular tweet.~\label{fig:CN_button}
\end{figure*}

If a user wants to monitor islamophobic discourse \textit{in real time}, the Twitter API is called on the fly searching for one of the key terms or hashtags described before, and several analyses are displayed. These include the 5 islamophobic keywords and hashtags that have been most recently used, the network of hashtags and keywords that are currently being used in association with the one of interest, as well as the most recent tweets posted online that contain an islamophobic term. A screenshot of this view is displayed in Figure
\ref{fig:CN_button}. Hashtag monitoring has been included in the platform as a straightforward way to track which semantic domains revolve around islamophobic messages and how they change over time. This is supported by the clustering functionality included in the platform based on Louvain algorithm \cite{2008JSMTE..10..008B}, which automatically identifies the hashtags and keywords that tend to co-occur and marks them in different colours. By monitoring hashtags it is possible to understand, for instance, whether some external events have an impact on online discourse against Muslims, and how islamophobic content is framed.

\begin{figure*}[h!]
  \centering
  \includegraphics[width=1\textwidth]{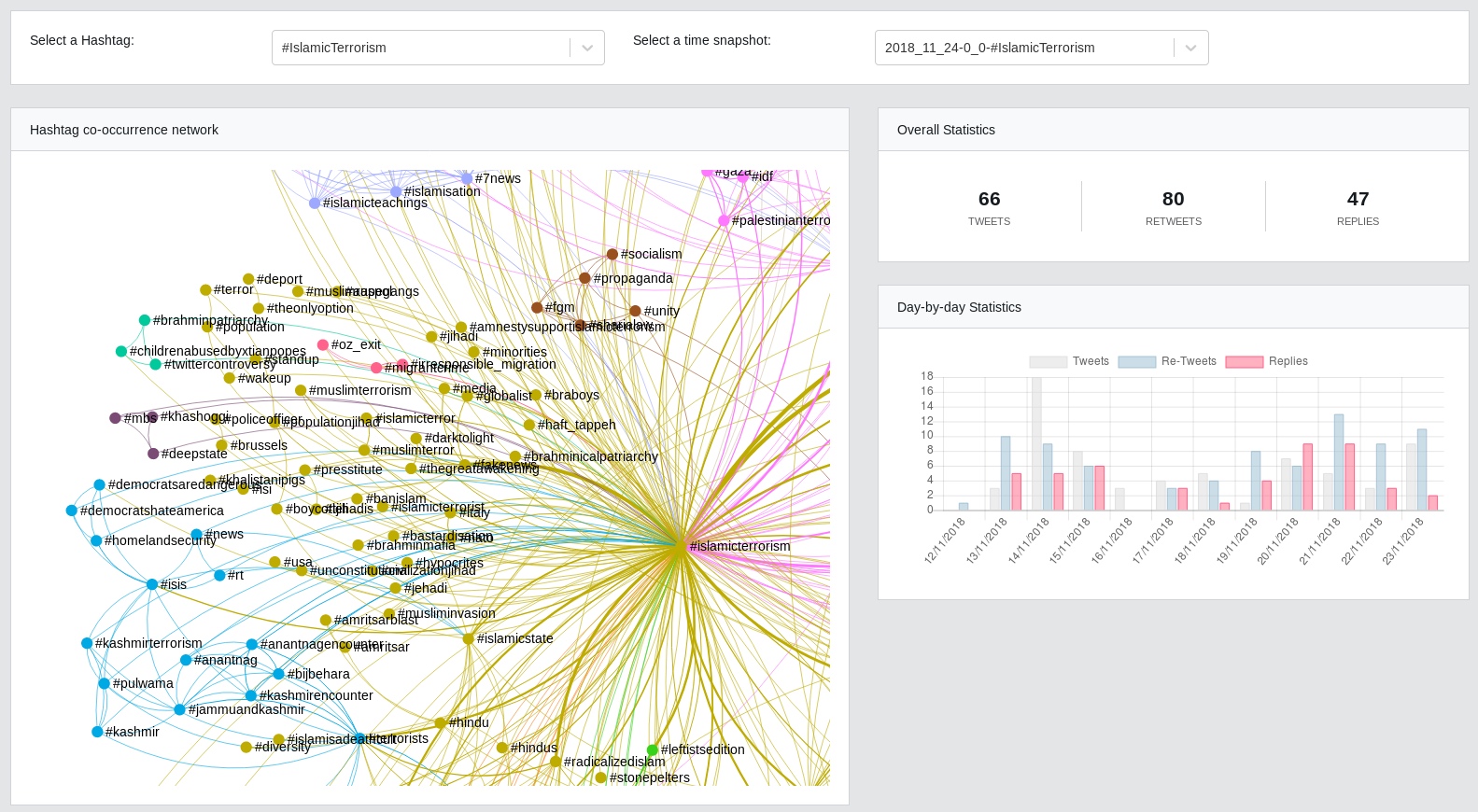}
  \caption{Screenshot of the monitoring view for past tweets: after selecting the \#IslamicTerrorism hashtag and the time slot, the network of co-occurring hashtags is displayed, together with overall statistics on the use of the hashtag in the period of interest.}~\label{fig:HashNet}
\end{figure*}

\begin{figure*}[h!]
  \centering
  \includegraphics[width=1\textwidth]{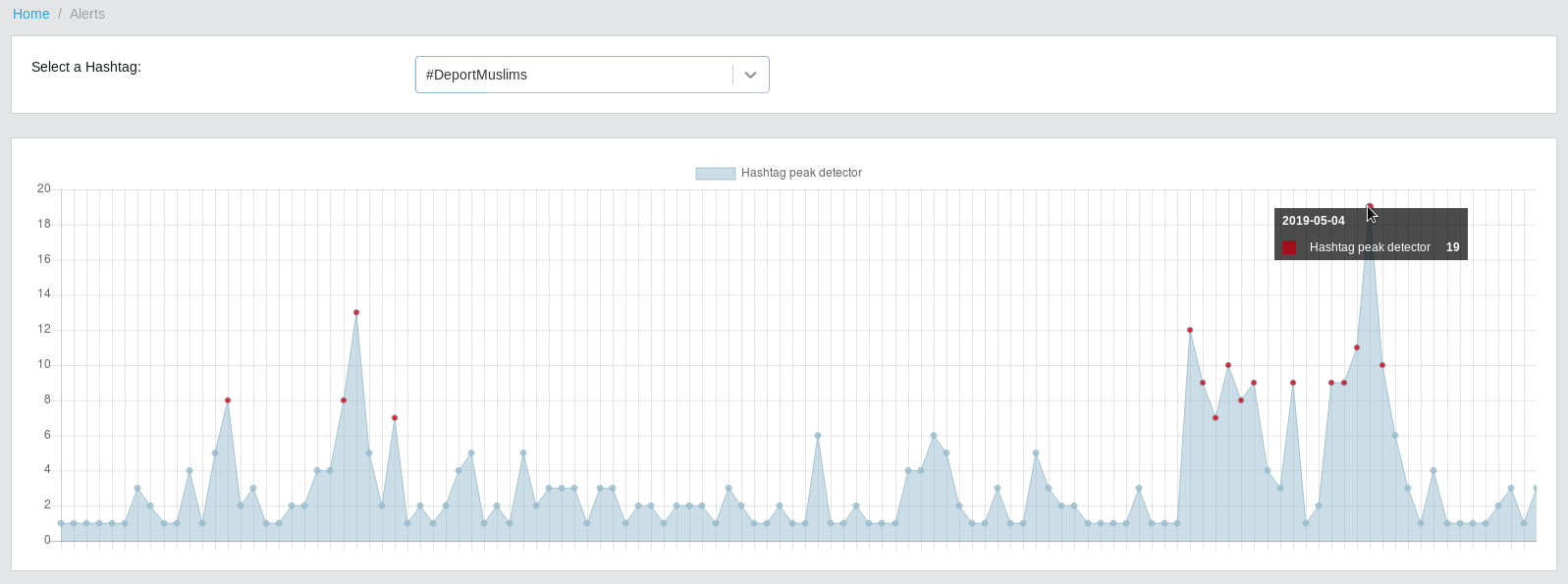}
  \caption{Screenshot of the peak detector view: after selecting the \#DeportMuslims hashtag, the number of occurrences for the given hashtag on each monitoring day is reported. Peaks are highlighted in red, with the associated date.}~\label{fig:Peaks}
\end{figure*}

The monitoring activity can also be \textit{retrospective}, i.e. concern islamophobic messages posted online in the past. The platform includes therefore the possibility to select a time window of 10 days and a predefined hashtag/keyword of interest, and to display the online activity around this term, such as the hashtags that most co-occurred with the one of interest (Fig. \ref{fig:HashNet}) and the most popular messages containing it. It is also possible to display the frequency with which the term has been posted online in the last months, with peaks flagged with a red dot when the frequency is above the average plus one standard deviation (Fig. \ref{fig:Peaks}), so that operators can use this information to connect peaks with past events occurring on specific dates.

\begin{figure*}[h!]
  \centering
  \includegraphics[width=1\textwidth]{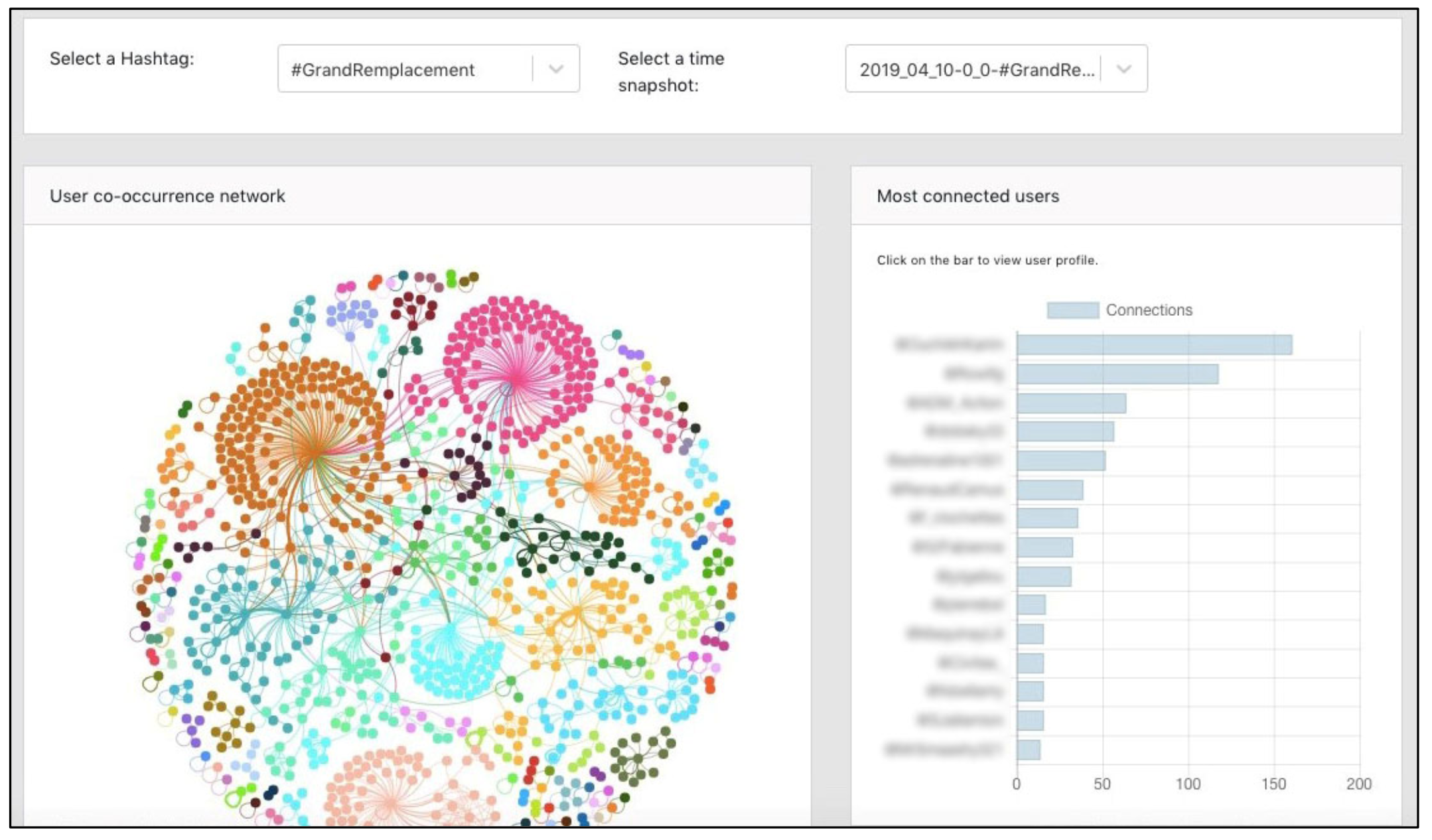}
  \caption{Screenshot of user monitoring view. Each node in the network is a user (names would become visible by zooming in). Names of most connected users have been blurred for privacy reasons.}~\label{fig:users}
\end{figure*}

Retrospective monitoring includes also the identification of the online communities that have shared islamophobic content in the selected time period. Starting from the domain-specific hashtags and keywords,  users that have either posted or retweeted the related messages are extracted and organised in a network, where users that tend to interact more often are represented as highly connected nodes and build sub-communities that the system displays in different colours (see Fig. \ref{fig:users}). By clicking on a node, representing a user, it is possible to see the user's number of tweets and retweets related to islamophobic content over time. There is also the ``Most Connected Users'' view, where the platform displays a ranked list of users with the most connections inside the network, i.e. those that are more likely to give visibility to Islamophobic messages. This information is used by NGO operators to understand whether the network sub-communities remain the same over time or whether they evolve, whether the groups of users are characterised by similar traits, such as political affiliation, and whether some users play a particularly relevant role in actively spreading anti-Muslim hatred. Although all displayed information is public, this kind of monitoring is highly sensitive, therefore the platform is password-protected and a personal account is given only to NGO operators involved in the activities against hate speech.

From a technical point of view, the platform relies on a relational database and a Tomcat application server. The interface is based on existing javascript libraries such as C3.js (\url{https://c3js.org}), D3.js (\url{https://d3js.org}) and Sigma.js (\url{http://sigmajs.org}).

\subsection{Counter-Narrative Suggestion Approach}
\begin{figure}[h!]
  \centering
  \includegraphics[width=1\columnwidth]{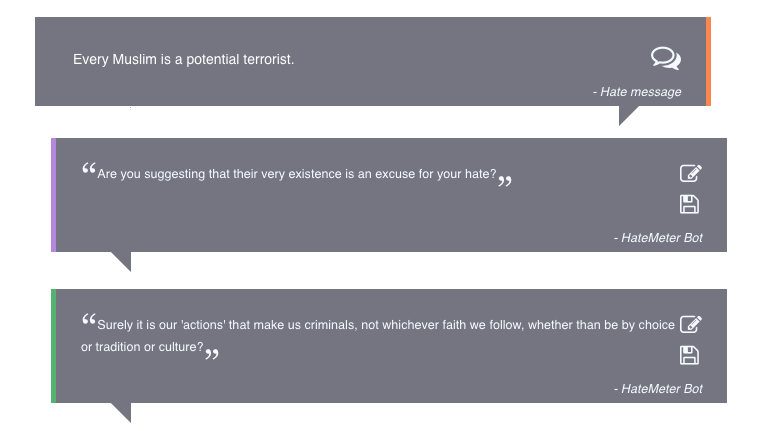}
  \caption{On top the hate content ``Every Muslim is a potential terrorist". Below a list of possible counter-narratives automatically suggested for that hate content.}~\label{fig:CN_suggestion}
\end{figure}

The main hate-countering feature consists in assisting operators in writing counter-narratives (CN) based on automatically generated suggestions. There is a specific tab of the interface that can be either activated from each hate tweet (see the blue bubble button present for each tweet in Figure \ref{fig:CN_button}) or via menu. While in the former case the content of the hate tweet and CN suggestions are directly displayed, in the latter case the user must copy-paste the hate content of interest. Figure \ref{fig:CN_suggestion} illustrates how the platform offers a list of suggestions for a given hate content. As displayed in the figure, if a user enters a hate message in the top box and clicks on the dialog icon on the right, the platform will output a list of counter-narrative suggestions below the hate content. For each suggestion, a user can directly modify/delete the suggestion if necessary or copy it to any social media platform for intervention. At the end of the page of the platform, a user can add a new counter-narrative from scratch if needed. In this regard, we can measure if the CN suggestion tool is useful in assisting writing a new counter narrative.

For implementing the counter-narrative suggestion tool, we have adopted a data-driven approach and we used an existing database \cite{chung2019conan} of hate message/counter-narratives written by NGO operators. The dataset consists of $\approx$6K hate message/counter-narrative pairs for English, $\approx$5K for French, and $\approx$3K for Italian, see \cite{chung2019conan} for further details on this dataset.

As a suggestion methodology, we have implemented a tf-idf information retrieval model \cite{Jones72astatistical} that, given a hate message as input, selects the most suitable responses (i.e. counter-narratives) from a database. tf-idf is a statistical measure that shows the importance of a term for a given document in a collection of documents and it is calculated as the product of the term frequency (tf) and the inverse document frequency (idf). 
We built our tf-idf model by calculating the tf-idf word-document matrix using the collected hate message/counter-narrative pairs as the document collection. For retrieving the counter-narrative suggestions, we first generated the tf-idf vector of the hate message given in input by an operator. Given the set of all hate message/counter-narrative pair vectors, the ones with the highest cosine similarity w.r.t. the input hate message vector are selected as the output pairs. Then, the counter-narratives of the top 4 pairs are presented in the platform together with an empty response text box in order to let operators write their own message if none of the suggestions presented is to their liking. Such new messages will be acquired as novel examples and included in our database.

We use a standard information retrieval approach since we aim to provide carefully created, grammatically and semantically correct responses to the operators. Although neural response generation \cite{shang2015neural,VinyalsL15} and argument generation \cite{le2018dave,hua2019argument} techniques have also been proposed recently, they still have limitations that might lead to substandard generations. For instance, the tendency to produce responses that are dull, generic and not engaging \cite{sordoni2015neural,serban2016building,li2016diversity}, or to overproduce negations \cite{hua2019argument} would be impractical to effectively counter online hatred. Furthermore, since deep neural models are inherently data-hungry, they need much more data than what we collected to scale up properly. From an  argumentation point of view, \cite{hua2019argument} discusses that automated generation systems have not been able to fully capture the organization and complexity of  human arguments yet. Considering the aforementioned issues, we adopt an information retrieval strategy, which guarantees informative, engaging and well-organized responses for our counter-narrative suggestion tool.

\section{Experiment Setup}

The goal of our experiment is to evaluate the effect of using our platform -- and in particular the suggestion tool -- in daily activities of NGO operators to fight hatred online. In our experiments we decided to focus on the CN aspect of the platform instead of monitoring. Since some of the monitoring functions are simply not doable by hand, it is not possible to obtain a clear control condition over them. On the contrary, the counter-narrative activity has always been carried out manually by operators and we can clearly measure the effect of its partial automation. In particular, we test if the CN suggestion approach is effective to help operators construct CNs in terms of quantity, usability (and quality).

\textbf{Design.} To test this effect we set up an evaluation framework with a between-subjects design and two conditions: using the platform with or without the counter-narrative suggestion tool (CN$^+$ and CN$^-$ conditions respectively). We summarize the tasks required for the CN$^+$ and CN$^-$ conditions in Table \ref{tab:task_overview}.

\textbf{Subjects.} In total 112 operators from three NGOs in different countries (UK, France, Italy) took part in the experiment. Specifically, \textbf{19} from UK, \textbf{42} from France, and \textbf{51} from Italy. These subjects are all volunteers that regularly contribute to NGO activities to fight islamophobia online and undergo specific training to comply with NGO code of conduct and fairness when it comes to online behaviour. Subjects were randomly assigned to one of the conditions, so that each operator was only exposed to a single interface version. In particular, 62 subjects were assigned to the CN$^+$ condition and 50 to the CN$^-$.

\textbf{Procedure.} For the experiment procedure, a deployment day was set up for the CN$^-$ version and another one for the CN$^+$ version of the platform. Operators were asked to gather in NGO's premises on the day of the evaluation, and everyone was equipped with a computer connected to the platform. The first part of the deployment day was meant to introduce the platform and its functionalities. After the introduction, a phase of some usability exercises/tasks was carried out to prime each operator in the same way and to ensure that they tried all the characteristics of the platform. Finally a questionnaire was administered to the subjects. For each language/NGO we followed the same procedure and a total of 6 deployment days (2 for each NGO) were thus organized.

\textbf{Instructions.} The usability tasks were meant to let operators inspect all the functionalities of the platform. Four dimensions of hate monitoring functionalities are included in the tasks: hate distribution, hate timeline, islamisation analysis and recent trends. For example islamisation analysis task was ``\textit{Starting from the `Hashtag trends' window, please analyse the key hashtags and keywords related to the concept of `Islamisation', checking the data obtained from the network analysis and the graphs that the platform accomplishes. Later, please identify the hate speakers who mostly utilize the hashtags and keywords that you have identified}". The instruction for each task is given in \ref{sec:usability_tasks_instruction}. For the CN$^+$ condition, we asked operators also to identify the most relevant tweets and use the suggestion tool to answer them, as they would have done in their normal CN activity. 

\textbf{Measurements.} For each condition we collected a mix of subjective (questionnaire) and objective (based on activity logs) measurements. Subjective responses were both quantitative and qualitative.

\section{Questionnaires}
To evaluate the operators' experience with the platform, two questionnaires were used: the User Experience Questionnaire (UEQ) \cite{laugwitz2008construction} and an additional questionnaire paired with some qualitative questions specifically developed for the hate countering task of our interface. 

\subsection{UEQ}
UEQ is a questionnaire designed to measure subjective user experience of interactive products. It covers 6 scales that evaluate attractiveness, hard usability aspects (perspicuity, efficiency, and dependability), and soft user experience aspects (stimulation and novelty). Each scale consists of several items with the form of semantic differential that is presented in 2 terms with opposite meanings. For each item, operators were asked to give a score between -3 and 3 on a 7-point Likert scale. The 6 scales used in UEQ are: 
\textsc{attractiveness}, general impression and aesthetics of the platform; 
\textsc{perspicuity}, the difficulty to familiarize the platform;
\textsc{efficiency}, the effort required to solve tasks;
\textsc{dependability}, the level of control and trust on the platform;
\textsc{stimulation}, the interest to use the platform; 
\textsc{novelty}, the novelty about the platform.

\subsection{Additional Questionnaire and Open-ended Questions}
In addition to the UEQ tool, we used rating scale questions and open-ended questions to examine the qualitative performance of the platform on specific aspects related to our task/scenario, specifically:

\begin{itemize}
\item \textsc{Monitor:} How satisfied are you with the reliability of the platform in monitoring anti-Muslim hatred online?
\item \textsc{Analysis:} How satisfied are you with the reliability of the platform in analysing anti-Muslim hatred online?
\item \textsc{Support:} How satisfied are you with the reliability of the platform in its ability to support NGOs in preventing and fighting online Islamophobia?
\item \textsc{Ease of use:} How satisfied are you with the platform's ease of use? 
\item \textsc{Look:} How satisfied are you with the look of the platform?
\end{itemize}

Moreover, in addition to above 5 questions, we asked one extra question, in the CN$^+$ condition, to evaluate the counter-narrative suggestion tool: 

\begin{itemize}

\item \textsc{Suggestion:} How satisfied are you with the counter-narrative suggestion tool?
\end{itemize}

For each question, operators were asked to give a score between 0 and 10 on a 10-point Likert scale and the reasons for assigning this score.

\begin{table}[h]
\caption{Overview of the Tasks for CN$^-$ and CN$^+$ conditions}
\label{tab:task_overview}
\centering
\begin{tabular}{|l|c|c|}
\hline
\textbf{Tasks}                                 & \textbf{CN$^-$} & \textbf{CN$^+$}  \\ \hline
\textit{Usability tasks}                       &     &                  \\ 
    \, \, Hate  distribution                   &  yes & yes             \\ 
    \, \, Hate  timeline                       &  yes & yes             \\
    \, \, Islamisation  analysis               &  yes & yes             \\
    \, \, Recent trends                        &  yes & yes             \\      \hline
\textit{CN writing}                            &  yes & yes             \\      \hline
\textit{Questionnaires}                        &      &                 \\      
    \, \, UEQ                                  &  yes & yes             \\ 
    \, \, Additional Questionnaire             &      &                 \\ 
    \, \, \, \, \, \, Monitor                  &  yes & yes             \\ 
    \, \, \, \, \, \, Analysis                 &  yes & yes             \\
    \, \, \, \, \, \, Support                  &  yes & yes             \\
    \, \, \, \, \, \, Ease of use              &  yes & yes             \\
    \, \, \, \, \, \, Look                     &  yes & yes             \\
    \, \, \, \, \, \, Suggestion               &  no  & yes             \\      \hline
\end{tabular}
\end{table}

\section{Data Analysis}
In this section, we first describe the outcome of the usability analysis for the platform according to the two questionnaires mentioned above. Then, we describe the analysis of the activity logs for the counter-narrative generation part.  

\subsection{UEQ Analysis}
As stated, the analysis was performed using the UEQ analysis tool\footnote{https://www.ueq-online.org}. We first performed inconsistency analysis and discarded inconsistent responses following the standard procedure indicated by the questionnaire. After inconsistency analysis, 9 responses were removed (2 responses from the CN$^+$ condition and 7 responses from CN$^-$).

In Table \ref{tab:UEQtable1}, we report the statistics of UEQ scales comparing the two conditions. The range of possible values is comprised between -3 (extremely negative) and 3 (extremely positive). According to UEQ tool, values between -0.8 and 0.8 correspond to a neutral evaluation, while values above 0.8 represent a positive and values below -0.8 express a negative evaluation. As can be observed in the table, all scales show a positive evaluation from users in both conditions, but the setting with the counter-narrative tool got a more positive evaluation. Among the scales, the one which received the worst judgements is Dependability, i.e. whether the user feels in control of the interaction. Possible explanations related to this aspect are reported in the open-ended questionnaire. The highest scores in both conditions concern the Attractiveness, i.e. overall impression of the platform, and the Stimulation, i.e. whether it is exciting or stimulating to use the system. 
In general, all values show that operators have a positive overall impression of the platform. This is rather expected, since previously monitoring analyses were performed manually by operators and many of the operations available with the platform can simply not be done with the standard social media search functions (e.g. hashtag trends going back to several months, or identification of sub-communities spreading islamophobic messages). 

\begin{table}
  \caption{UEQ Scales with and without Counter-narrative Suggestion Tool (CN$^+$ and CN$^-$ respectively). CN$^{-/+}$ denotes CN$^-$ receiving suggestion tool in a post-hoc experiment}
  \label{tab:UEQtable1}
  \small
  \centering
  \begin{tabular}{l r r r r r r}
     & \multicolumn{2}{c}{\small{\textbf{CN$^-$}}} &  \multicolumn{2}{c}{\small{\textbf{CN$^+$}}}
     & \multicolumn{2}{c}{\small{\textbf{CN$^{-/+}$}}}
     \\
    \cline{2-3} \cline{4-5} \cline{6-7}
    {\small\textit{Scale}}
      & {\small \textit{Mean}}
    & {\small \textit{SD}}
    & {\small \textit{Mean}}
    & {\small \textit{SD}} 
    & {\small \textit{Mean}}
    & {\small \textit{SD}}\\
    \hline
    Attractiveness &  1.62 & 0.77 & 1.74 & 0.70 & 2.11 & 0.28 \\
    Perspicuity & 1.21 & 1.14 & 1.33 & 0.75  & 1.36 & 0.49\\
    Efficiency & 1.31 & 0.85 & 1.47 & 0.73  & 1.69 & 0.68\\
    Dependability & 0.98 & 0.71 & 1.17 & 0.52 & 1.11 & 0.47 \\
    Stimulation & 1.58 & 0.84 & 1.81 & 0.74  & 1.78 & 0.32\\
    Novelty & 1.27 & 1.17 & 1.75 & 0.67 & 1.83 & 0.42 \\
    \hline
  \end{tabular}
\end{table}

With regards to counter-narratives, operators reported higher evaluation for the condition where the suggested responses provided by the CN suggestion tool is present, with Novelty and Stimulation scales showing most improvements. Users may feel stimulated to adopt the CN tool especially when no suitable counter-narrative idea is available to operators or time is a concern. On the other hand, one limitation of the automated suggestion mechanism is that the collected responses (i.e. CNs written by operators in our experiments) may fall within the scope of suggested responses. The collected responses can be lexically or semantically similar to the suggested counter-narratives, since operators were primed by given counter-narrative references. Hence, they are more likely to think and write towards the reference provided. In Section \ref{sec:limitation} we will further discuss this issue and possible mechanisms ensuring and controlling the diversity in counter-narratives collected.

The UEQ scales can be further grouped into three categories: Attractiveness, Pragmatic quality (including Perspicuity, Efficiency, Dependability) and Hedonic quality (including Stimulation and Originality), where pragmatic quality assesses goal-directed quality aspects and hedonic quality measures quality aspects that are not related to the platform goal of monitoring and countering hate speech. In Table \ref{tab:PragmaticTable}, we provide the mean of the three pragmatic and hedonic quality aspects. As can be observed in the Table, all scales receive scores higher than 1.1, and the condition with CN is better evaluated than the one without CN for all the categories. 

\begin{table}
\caption{Pragmatic and Hedonic Quality}\label{tab:PragmaticTable}
  \centering
  \small
  \begin{tabular}{l r r r}
     & \small{\textbf{CN$^-$}} & \small{\textbf{CN$^+$}} &
     \small{\textbf{CN$^{-/+}$}} \\
    \cline{2-4} 
    {\small\textit{Scale}} & {\small \textit{Mean}} & {\small \textit{Mean}} & {\small \textit{Mean}} \\
    \hline      
    Attractiveness &  1.62 & 1.74 & 2.11 \\
    Pragmatic Quality & 1.17 & 1.32 & 1.39 \\
    Hedonic Quality & 1.43 & 1.78 & 1.81 \\
    \hline     
  \end{tabular}
\end{table}

We further provide a comparison between our platform and the existing benchmark for the UEQ tool to better understand how our UEQ scores compared to those of other products. The benchmark dataset involves 18,483 participants judging 401 products, including web pages, social media, and business software. As shown in Table \ref{tab:benchmark}, the benchmark categorizes a product into five groups: (1) excellent, which falls within the first 10\% results, (2) good, where 10\% of the results in the benchmark are better than the result for the evaluated product, while 75\% of the results are worse, (3) above average, where 25 \% of the results in the benchmark are better than the result for the evaluated product, while 50\% of the results are worse, (4) below average, where 50 \% of the results in the benchmark are better than the result for the evaluated product while 25\% of the results are worse, and (5) bad, which is among the last 25 \% results. 

\begin{table}[h]
\caption{Bechmark Intervals for the UEQ Scores (per Scale)}
\label{tab:benchmark}
\centering
\begin{tabular}{|l|c|}
\hline
\textbf{categories} & \textbf{evaluation results}  \\ \hline
excellent           &  90\% $<$ results $\leq$ 100\%        \\ \hline
good                & 75\% $<$ results $\leq$ 90\%          \\ \hline
above average       & 50\% $<$ results $\leq$ 75\%          \\ \hline
below average       & 25\% $<$ results $\leq$ 50\%          \\ \hline
bad                 & results $\leq$ 25\%                   \\ \hline
\end{tabular}
\end{table}

\begin{figure*}
  \centering
  \includegraphics[width=1.8\columnwidth]{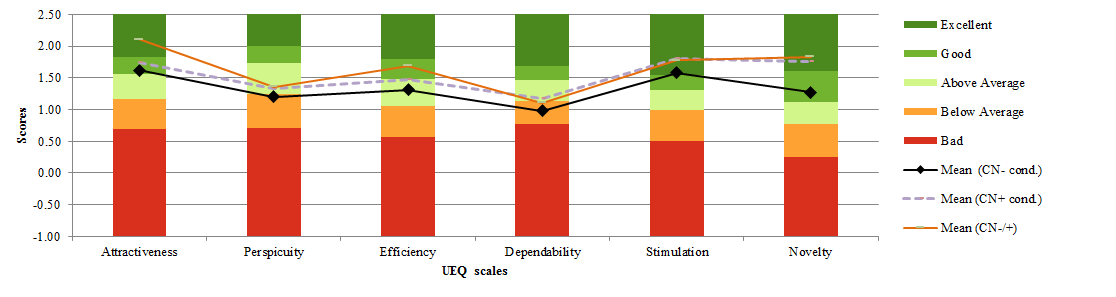}
  \caption{Statistics on CN$^-$ condition, CN$^+$ condition and CN$^{-/+}$, compared with the benchmark for UEQ scales.}~\label{fig:benchmark}
\end{figure*}

The benchmark chart is displayed in Figure \ref{fig:benchmark}. The analysis shows that all scales for our platform without suggestion tool (CN$^-$ cond in the chart) achieve above average performance except for Perspicuity and Dependability. The latter scale in particular shows that the platform without counter-narrative functionalities supports operators in their activity only to a limited extent. This is however mitigated in the CN$^+$ condition, achieving an evaluation above the average for all scales, and even an excellent evaluation in the Attractiveness, Stimulation and Novelty scales. This suggests that users consider the counter-narrative tool to be innovative and creative, and that they prefer the platform including the suggestion engine than the one without it.

To further validate this finding, we run a post hoc experiment with 10 operators (randomly selected) by replicating the experiment also in a within-subject design (i.e. after few weeks, some operators that were assigned to the CN$^-$ condition were administered also with the CN$^+$ condition). With this experiment, that we indicate with CN$^{-/+}$, we tested if decisions under separate evaluation diverge from decision under joint evaluation. In this way, we controlled for the possible problem of insensitivity \cite{kahneman1992valuing,desvousges1993measuring} that tends to affect specifically between subject design and we weighed the potential difficulty that operators have in accurately perceiving interface differences in a between design (operators in CN$^-$ condition did not have to write counter-narratives during testing), by forcing individuals to experience scope differences in a within design. The results of this second evaluation are also reported in Tables \ref{tab:UEQtable1} and \ref{tab:PragmaticTable}, and also in Figure ~\ref{fig:benchmark}. Results show, as expected, that for most scales the differences with respect to CN$^-$ condition are more evident: when possible masking effects of novelty and insensitivity are controlled, the advantages of integrating the CN suggestion tool appear even more clearly.

\subsection{Additional Questionnaire and Open Questions Analysis}
We performed an additional qualitative evaluation to assess how the platform can be used to monitor online islamophobia and support NGOs to fight anti-Muslim speech, using rating scale questions and an open-ended questionnaire. The results for rating scale questions, discussed in the previous section, are reported in Table \ref{tab:indicator_tab1e}. Overall, the platform received positive reviews on all dimensions considered in the evaluation, with at least a median value of 7 and a mean value of 6.8. Regarding the suggestion tool condition, users reported higher ratings overall in contrast to a scenario where no suggestion tool is available, except for a slightly lower value concerning ease of use. Also in this case the results from CN$^{-/+}$ configuration show that the differences with respect to CN$^-$ tend to increase. 

Explanations of the reasons behind this evaluation and additional comments have been provided by operators in the replies to the open-ended questionnaire, which we summarize below.

\begin{table*}[t]
\caption{Statistics of 6 Indicators from Additional Questionnaire on CN$^-$, CN$^+$ and CN$^{-/+}$ Conditions}\label{tab:indicator_tab1e}
  \small
  \centering
  \begin{tabular}{l r r r r r r r r r r r r}
    & \multicolumn{4}{c}{\small{\textbf{CN$^-$}}} &  \multicolumn{4}{c}{\small{\textbf{CN$^+$}}}
    &  \multicolumn{4}{c}{\small{\textbf{CN$^{-/+}$}}} \\
    \cline{2-13}
     & {\small\textit{Mean}} &  {\small\textit{Median}} &  {\small\textit{SD}} & {\small\textit{N}} & {\small\textit{Mean}} &  {\small\textit{Median}} &  {\small\textit{SD}} & {\small\textit{N}} & {\small\textit{Mean}} &  {\small\textit{Median}} &  {\small\textit{SD}} & {\small\textit{N}}\\
    \hline      
    \small{Monitor} & 7.33 & 8.00 & 1.62 & 55 & 7.48 & 8.00 & 1.56 & 49 & 7.78 & 8.00 & 0.97 & 9 \\
    \small{Analysis} & 6.84 & 7.00 & 2.03 & 55 & 7.29 & 7.00 & 1.47 & 49 & 7.33 & 8.00 & 1.41 & 9\\
   \small{Support} & 7.33 & 8.00 & 1.88 & 55 & 7.46 & 8.00 & 1.86 & 49 & 8.00 & 8.00 & 0.87 & 9 \\
   \small{Ease of use} & 7.51 & 8.00 & 1.93 & 55 & 7.35 & 8.00 & 1.87 & 49 & 7.78 & 8.00 & 1.48 & 9 \\
   \small{Look} & 7.51 & 8.00 & 1.88 & 55 & 7.60 & 8.00 & 1.97 & 49 & 8.00 & 8.00 & 1.22 & 9 \\
   \small{Suggest} & - & - & - & - & 6.85 & 7.00 & 2.06 & 49 & 6.89 & 7.00 & 2.15 & 9 \\ 
    \hline      
  \end{tabular}
\end{table*}

\subsubsection{Monitor}
For all the three languages, users generally concluded that the platform is useful and reliable in monitoring anti-Muslim tweets. As expected, no remarkable differences in the comments can be found between CN$^-$ and CN$^+$ conditions. The most appreciated features are the possibility to spot spikes in the hashtag frequency and to show users divided into sub-communities. Some operators are not completely satisfied with the choice to focus only on Twitter, and with the hashtag-based monitoring process, arguing that islamophobic messages that do not contain one of the predefined hashtags are not included in our view.

\subsubsection{Analysis}
Users commented that the platform overall is innovative and useful for analysing anti-Muslim hatred. In particular, diagrams are deemed useful in providing very clear analyses and are easy to interact with. Some operators would like to have more user analysis features, in order to profile specific haters and monitor them. However, this task poses ethical questions when it is done at scale, and this is also forbidden by the European General Data Protection Regulation 2016/679.\footnote{\url{https://eur-lex.europa.eu/legal-content/EN/TXT/?uri=CELEX\%3A32016R0679}} Like in the \textit{Monitor} evaluation, there is no difference in the comments between CN$^-$ and CN$^+$ condition because both the analysis and the monitoring tasks are not among the goals of the counter-narrative engine.

\subsubsection{Support}
Comments related to the ability of the platform to support NGOs in preventing and fighting Islamophobia online are overall positive, although operators tend to specify that this is just an impression and that they would need more time to test the effectiveness of the platform in the long run. Most operators appreciate the novelty of the approach compared to the manual one, since aggregated data provide an added knowledge on Islamophobic trends that the monitoring of single users or hashtags does not cover. In some comments, they suggest introducing the possibility to export such data for preparing reports and supporting long-term analyses.
In the CN$^+$ condition, some comments concern the counter-narrative tool, which is deemed helpful since it is often difficult to form new responses and critical judgements from scratch. Furthermore, with the suggestion tool operators can decrease time and effort spent on writing the responses. 

\subsubsection{Ease of use}
As for the ease of use, participants commented that the experience in using the platform is very intuitive and user-friendly. They appreciated in particular the fact that complex analyses were made understandable in an easy way. Few suggested to add a \emph{help} function where more details are provided about how the displayed analysis was exactly extracted. In the CN$^+$ condition, the comments were similar to the ones in the CN$^-$ condition, apart from one that mentioned explicitly that the counter-narrative tool was intuitive to use.

\subsubsection{Look}
While some users reported that the aesthetics of the platform is not as important as its usage and functionality, the overall evaluation was positive. The use of colours was appreciated by the operators, and some suggested that they could be used even more to guide the interpretation of the results. In the CN$^+$ condition, no specific comment concerned the counter-narrative interface alone. 

\subsubsection{Suggestion}
The question about the counter-narrative suggestion tool was present only in the CN$^+$ condition. The evaluation by operators was very positive, with some comments stating that this is the most useful feature in the platform and that it could be a game changer in fighting islamophobia online, especially for operators who are not experienced and have to undergo training on counter-narrative strategies. Some comments state that the provided answers should be rather used as a template or a suggestion, and not be posted verbatim. Nevertheless, operators believe that this would be less time consuming than writing counter-narratives from scratch.

\subsection{Content Analysis and Activity Logs}
Since previous evaluations suggest that using the counter-narrative tool may be less time-consuming than manually writing messages from scratch, we added a final evaluation to measure this aspect. For this reason, we kept the logs of operator activities during the deployment days, and in particular we recorded the time needed in writing new responses per CN suggestion. The assumption is that with the CN suggestion tool the time needed for composing new or modifying responses would be shorter compared to the one without suggestion tool.

Therefore we recorded how many suggested messages were further modified and used by operators, and how many were written by operators and saved through the platform after seeing the CN suggestions. In total, using the platform we collected 999 modified messages and 335 new counter-narratives across three countries. This implies that our suggestion tool based on tf-idf has around 75\% accuracy, i.e. in three cases out of four it was helping operators in their countering activity by providing relevant suggestions. If we focus on relevance of suggestions, using the cosine similarity between actual HS and suggested CNs, we found that the score of modified suggestions is almost double than the similarity of suggestions that have been discarded (0.13, 0.12, 0.06 vs. 0.07, 0.07, 0.03 for EN, IT, FR respectively). This implies that there is definitely a relevance phenomenon in the selection process. 
Finally, as it is shown in Table \ref{tab:time_table1}, for the three data collections there is expectedly a significant time decrease by half after the suggestion tool is introduced. The numbers indicate the average time required to complete the task (including reading CN suggestions, for the CN$^+$ condition).

For CN$^-$ condition, we estimated the time needed for writing a CN from scratch without the CN suggestion tool in a separate session for each country. We divided the total recorded time by the total number of CNs collected, resulting in on average 8 minutes per CN, which is similar to the one reported in prior study on counter-narrative data collection through NGO operators \cite{chung2019conan}. This time includes  possible searching for useful information on the web and structuring a proper argument before writing the corresponding counter narrative.

For the CN$^+$ condition, the time needed to modify a suggestion or to write a new CN is similar\footnote{The suggested CNs in the database are generally short and easily understandable with 2 to 3 sentences, so the time for reading them is almost negligible as compared to the time for writing or searching for useful information.}. Since operators in CN$^+$ condition composed new CNs after seeing suggestions, we hypothesised that  these ``prompts" reduced the time required to search for useful information and digest it as compared to CN$^-$ condition. We performed also an additional analysis, computing the Human-targeted Translation Edit Rate, aka HTER \cite{Specia10AMTA}, for the modified CNs. HTER is a measure primarily used in Machine Translation to compute the post-editing effort of annotators for correcting the output of automatic MT systems. We found that the average HTER is above 0.4 for all languages, which represents an empirical threshold above which correction or rewriting from scratch are almost the same \cite{turchi-etal-2013-coping}. This explains why the two conditions have similar timing.
This result indicates that the suggestion tool we provided is able to offer suitable counter-narrative candidates and meets the goal of increasing efficiency of the operators. This efficiency is primarily obtained by providing hints and arguments to operators rather than simply allowing them to save writing time. In fact, CN$^+$ condition is always better than CN$^-$ for both modified and new CN configurations with almost comparable time. In fact, with this design, operators can utilize their time to refine responses on top of suggestions and construct more diverse counter-narratives. Considering the different countries, we observe that for both France and Italy, operators spent less time on modification than on writing new counter-narratives. This implies that our suggestion tool is of good quality and can provide proper counter-narratives after small adaptation. The fact that operators from UK spent slightly less time writing new counter-narratives than adapting the suggested ones may depend on the fact that this user group was the only one including some professional operators, and in general volunteers from this specific NGO receive more training than the others. This may explain why UK operators need to personalise more the suggested counter-narratives, and are faster in writing their own. 

\begin{table}[h!]
  \caption{Time required for writing CN (Modified CN based on suggestions vs. New CN from scratch) across 3 Countries. Time is expressed in minutes}\label{tab:time_table1}
  \centering
  \small
  \begin{tabular}{l r r r r r r}
    & \multicolumn{3}{c}{\small{\textbf{CN$^-$ cond.}}} &  \multicolumn{3}{c}{\small{\textbf{CN$^+$ cond.}}} \\
    \cline{2-7}  
    & {\small \textit{UK}}
    & {\small \textit{FR}}
    & {\small \textit{IT}}
      & {\small \textit{UK}}
    & {\small \textit{FR}}
    & {\small \textit{IT}} \\
    \hline 
    modified\_CN$_\mu$  & -& -& -& 4.95 & 2.13 & 4.30\\
    new\_CN$_\mu$  & 8 & 8 & 8 & 4.21 & 3.62 & 4.98 \\
    \hline      
  \end{tabular}
\end{table}

\section{Limitations} \label{sec:limitation}
This work presents some limitations that we will possibly try to address in the future. First of all, we chose to build our counter-narrative tool around a retrieval-based suggestion engine, that produces fixed suggestions for similar hate messages. In this regard, users were instructed to paraphrase the suggested counter-narratives in order to increase the variability of the messages posted online. Still, the same themes/ideas in the suggestions may be re-used reducing the diversity of counter narratives. Several diversity metrics can be introduced to select more diverse suggestions with regards to the ones that are already employed by users, for instance, Jaccard similarity \cite{wang-2018-sketching}, Self-BLEU \cite{10.1145/3209978.3210080}, and Shannon
entropy \cite{manning1999foundations} based on n-grams. It may be possible to adopt an approach generating more diverse counter-narratives by taking advantage of recent advances in neural text generation \cite{radford2019language}. However, with such approaches it is difficult to control the fluency and the grammaticality of the generated messages, and we avoided them in the first experiments because badly written sentences may discourage operators from using the tool, or lead them to correct mistakes instead of focusing on the best counter-narrative strategy. As a next step, we will compare the quality of the suggested messages using a retrieval-based and a neural generation approach, measuring also the time spent to paraphrase or correct a counter-narrative.

Second, while we believe that the findings on the counter-narrative suggestion tool could apply to hate speech in general, we tested it only on the specific topic of islamophobia, because it was part of the planned activities of the three NGOs involved in the evaluation. On this topic, it was rather straightforward to identify hashtags to be monitored, as well as to collect examples of hate messages and counter-narratives. It is possible that hate messages with a different target, for example misogynistic ones, are less associated with hashtags, or are expressed in the more subtle form of microaggressions. It may, therefore, be more difficult to implement an easy yet effective counter-narrative strategy like the one we introduce in this paper.

Third, while results obtained by testing with 112 operators on 6 different deployment days demonstrate that our platform is helpful for assisting in monitoring and countering online hate messages, we recognize that a long-term evaluation is required through daily activities, to ensure the effectiveness and usability of the platform. As future work, we will extend the assessment period and evaluate the actual activity at scale.

\section{Conclusions}
As digital technologies have been deeply woven into our social lives and hate content spreads rapidly, approaches to deal with this issue at scale are very much needed, especially by organisations that fight hate content online but rely on a limited number of volunteers and operators. To this purpose, we have developed an online platform that monitors islamophobic messages on Twitter and supports NGO operators in quickly responding to hate messages with the help of a counter-narrative tool. 
We evaluated in particular the contribution of the counter-narrative suggestion tool for English, French, and Italian in addressing islamophobic messages. 
We performed our evaluation involving more than 100 NGO operators from three countries, using two questionnaires, a set of open-ended questions and analysing the user logs. 
The evaluation shows that operators have a positive attitude towards the adoption of our platform, finding it both effective and easy-to-use. The counter-narrative tool proved to efficiently support operators in responding to hate messages, also reducing the time needed for the task. Our work gives new insight into the adoption of digital tools to empower NGOs in countering online hate messages.    

\section*{Acknowledgments}
This work was partly supported by the HATEMETER project within the EU Rights, Equality and Citizenship Programme 2014-2020.  
We are grateful to the following NGOs and all annotators for their help: Stop Hate UK, Collectif Contre l'Islamophobie en France, Amnesty International (Italian Section - Task force  hate speech).

\bibliographystyle{plain} 
\bibliography{reference}

\begin{thebibliography}{10}

\bibitem{aguilera2016islamonausea}
Carmen Aguilera-Carnerero and Abdul~Halik Azeez.
\newblock `islamonausea, not islamophobia': The many faces of cyber hate
  speech.
\newblock {\em Journal of Arab \& Muslim Media Research}, 9(1):21--40, 2016.

\bibitem{awan2016islamophobia}
Imran Awan.
\newblock Islamophobia on social media: A qualitative analysis of the
  facebook's walls of hate.
\newblock {\em International Journal of Cyber Criminology}, 10(1), 2016.

\bibitem{bartlett2015counter}
Jamie Bartlett and Alex Krasodomski-Jones.
\newblock Counter-speech examining content that challenges extremism online.
\newblock {\em DEMOS, October}, 2015.

\bibitem{basile-etal-2019-semeval}
Valerio Basile, Cristina Bosco, Elisabetta Fersini, Debora Nozza, Viviana
  Patti, Francisco~Manuel Rangel~Pardo, Paolo Rosso, and Manuela Sanguinetti.
\newblock {S}em{E}val-2019 task 5: Multilingual detection of hate speech
  against immigrants and women in twitter.
\newblock In {\em Proceedings of the 13th International Workshop on Semantic
  Evaluation}, pages 54--63, Minneapolis, Minnesota, USA, June 2019.
  Association for Computational Linguistics.

\bibitem{quteprints101369}
Anat Ben-David and Ariadna Matamoros-Fernandez.
\newblock Hate speech and covert discrimination on social media: Monitoring the
  facebook pages of extreme-right political parties in spain.
\newblock {\em International Journal of Communication}, 10:1167--1193, 2016.

\bibitem{benesch2014countering}
Susan Benesch.
\newblock Countering dangerous speech: New ideas for genocide prevention.
\newblock {\em Washington, DC: United States Holocaust Memorial Museum}, 2014.

\bibitem{dangspe}
Susan Benesch, Cathy Buerger, Tonei Glavinic, and Sean Manion.
\newblock Dangerous speech: A practical guide.
\newblock {\em Dangerous Speech Project}, 2018.

\bibitem{benesch2016}
Susan Benesch, Derek Ruths, Kelly~P Dillon, Haji~Mohammad Saleem, and Lucas
  Wright.
\newblock Counterspeech on twitter: A field study.
\newblock {\em Dangerous Speech Project}, 2016.

\bibitem{2008JSMTE..10..008B}
Vincent~D. {Blondel}, Jean-Loup {Guillaume}, Renaud {Lambiotte}, and Etienne
  {Lefebvre}.
\newblock {Fast unfolding of communities in large networks}.
\newblock {\em Journal of Statistical Mechanics: Theory and Experiment},
  2008(10):10008, October 2008.

\bibitem{doi:10.1002/poi3.85}
Pete Burnap and Matthew~L. Williams.
\newblock Cyber hate speech on twitter: An application of machine
  classification and statistical modeling for policy and decision making.
\newblock {\em Policy \& Internet}, 7(2):223--242, 2015.

\bibitem{burnap2015cosmos}
Peter Burnap, Omer Rana, Matthew Williams, William Housley, Adam Edwards,
  Jeffrey Morgan, Luke Sloan, and Javier Conejero.
\newblock Cosmos: Towards an integrated and scalable service for analysing
  social media on demand.
\newblock {\em International Journal of Parallel, Emergent and Distributed
  Systems}, 30(2):80--100, 2015.

\bibitem{Capozzi:2018:DVP:3240431.3240437}
Arthur T.~E. Capozzi, Viviana Patti, Giancarlo Ruffo, and Cristina Bosco.
\newblock A data viz platform as a support to study, analyze and understand the
  hate speech phenomenon.
\newblock In {\em Proceedings of the 2Nd International Conference on Web
  Studies}, WS.2 2018, pages 28--35, New York, NY, USA, 2018. ACM.

\bibitem{capozzi2019computational}
Arthur~TE Capozzi, Mirko Lai, Valerio Basile, Fabio Poletto, Manuela
  Sanguinetti, Cristina Bosco, Viviana Patti, Giancarlo Ruffo, Cataldo Musto,
  Marco Polignano, et~al.
\newblock Computational linguistics against hate: Hate speech detection and
  visualization on social media in the" contro l’odio" project.
\newblock In {\em 6th Italian Conference on Computational Linguistics, CLiC-it
  2019}, volume 2481, pages 1--6. CEUR-WS, 2019.

\bibitem{chung2019conan}
Yi-Ling Chung, Elizaveta Kuzmenko, Serra~Sinem Tekiroglu, and Marco Guerini.
\newblock Conan-{CO}unter {NA}rratives through {N}ichesourcing: a multilingual
  dataset of responses to fight online hate speech.
\newblock In {\em Proceedings of the 57th Conference of the Association for
  Computational Linguistics}, pages 2819--2829, 2019.

\bibitem{chung2020italian}
Yi-Ling Chung, Serra~Sinem Tekiroglu, and Marco Guerini.
\newblock Italian counter narrative generation to fight online hate speech.
\newblock In {\em Proceedings of the Seventh Italian Conference on
  Computational Linguistics}, Bologna, Italy, 2020.

\bibitem{dang2018but}
Brandon Dang, Martin~J Riedl, and Matthew Lease.
\newblock But who protects the moderators? the case of crowdsourced image
  moderation.
\newblock {\em arXiv preprint arXiv:1804.10999}, 2018.

\bibitem{desvousges1993measuring}
William~H. Desvousges, F.~Reed~Johnson, Richard~W. Dunford, K.~Nicole~Wilson,
  and Kevin~J. Boyle.
\newblock Measuring natural resource damages with contingent valuation.
\newblock In {\em Contingent valuation: A critical assessment}, pages 91--164.
  Emerald Group Publishing Limited, 1993.

\bibitem{ernst2017hate}
Julian Ernst, Josephine~B Schmitt, Diana Rieger, Ann~Kristin Beier, Peter
  Vorderer, Gary Bente, and Hans-Joachim Roth.
\newblock Hate beneath the counter speech? a qualitative content analysis of
  user comments on youtube related to counter speech videos.
\newblock {\em Journal for Deradicalization}, pages 1--49, 2017.

\bibitem{gagliardone2015countering}
Iginio Gagliardone, Danit Gal, Thiago Alves, and Gabriela Martinez.
\newblock {\em Countering online hate speech}.
\newblock Unesco Publishing, 2015.

\bibitem{hall2013hate}
Nathan Hall.
\newblock {\em Hate crime}.
\newblock Willan, 2013.

\bibitem{hua2019argument}
Xinyu Hua, Zhe Hu, and Lu~Wang.
\newblock Argument generation with retrieval, planning, and realization.
\newblock {\em arXiv preprint arXiv:1906.03717}, 2019.

\bibitem{Jones72astatistical}
Karen~Sp\"{a}rck Jones.
\newblock A statistical interpretation of term specificity and its application
  in retrieval.
\newblock {\em Journal of Documentation}, 28:11--21, 1972.

\bibitem{kahneman1992valuing}
Daniel Kahneman and Jack~L Knetsch.
\newblock Valuing public goods: the purchase of moral satisfaction.
\newblock {\em Journal of environmental economics and management},
  22(1):57--70, 1992.

\bibitem{karunakaran2019testing}
Sowmya Karunakaran and Rashmi Ramakrishan.
\newblock Testing stylistic interventions to reduce emotional impact of content
  moderation workers.
\newblock In {\em Proceedings of the AAAI Conference on Human Computation and
  Crowdsourcing}, volume 7.1, pages 50--58, 2019.

\bibitem{laugwitz2008construction}
Bettina Laugwitz, Theo Held, and Martin Schrepp.
\newblock Construction and evaluation of a user experience questionnaire.
\newblock In {\em Symposium of the Austrian HCI and Usability Engineering
  Group}, pages 63--76. Springer, 2008.

\bibitem{le2018dave}
Dieu-Thu Le, Cam~Tu Nguyen, and Kim~Anh Nguyen.
\newblock Dave the debater: a retrieval-based and generative argumentative
  dialogue agent.
\newblock In {\em Proceedings of the 5th Workshop on Argument Mining}, pages
  121--130, 2018.

\bibitem{li2016diversity}
Jiwei Li, Michel Galley, Chris Brockett, Jianfeng Gao, and Bill Dolan.
\newblock A diversity-promoting objective function for neural conversation
  models.
\newblock In {\em Proceedings of NAACL-HLT}, pages 110--119, 2016.

\bibitem{manning1999foundations}
Christopher Manning and Hinrich Schutze.
\newblock {\em Foundations of statistical natural language processing (2nd
  printing)}.
\newblock MIT press, 2000.

\bibitem{mathew2018analyzing}
Binny Mathew, Navish Kumar, Pawan Goyal, Animesh Mukherjee, et~al.
\newblock Analyzing the hate and counter speech accounts on twitter.
\newblock {\em arXiv preprint arXiv:1812.02712}, 2018.

\bibitem{mathew2018thou}
Binny Mathew, Hardik Tharad, Subham Rajgaria, Prajwal Singhania, Suman~Kalyan
  Maity, Pawan Goyal, and Animesh Mukherje.
\newblock Thou shalt not hate: Countering online hate speech.
\newblock {\em arXiv preprint arXiv:1808.04409}, 2018.

\bibitem{menini2019system}
Stefano Menini, Giovanni Moretti, Michele Corazza, Elena Cabrio, Sara Tonelli,
  and Serena Villata.
\newblock A system to monitor cyberbullying based on message classification and
  social network analysis.
\newblock In {\em Proceedings of the Third Workshop on Abusive Language
  Online}, pages 105--110, 2019.

\bibitem{Mondal:2017:MSH:3078714.3078723}
Mainack Mondal, Leandro~Ara\'{u}jo Silva, and Fabr\'{\i}cio Benevenuto.
\newblock A measurement study of hate speech in social media.
\newblock In {\em Proceedings of the 28th ACM Conference on Hypertext and
  Social Media}, HT '17, pages 85--94, New York, NY, USA, 2017. ACM.

\bibitem{moretti-13}
Franco Moretti.
\newblock {\em {Distant Reading}}.
\newblock Verso, London, 2013.

\bibitem{munger2017tweetment}
Kevin Munger.
\newblock Tweetment effects on the tweeted: Experimentally reducing racist
  harassment.
\newblock {\em Political Behavior}, 39(3):629--649, 2017.

\bibitem{mandola}
D.~Paschalides, D.~Stefanidis, M.~Hernando, G.~Pallis, and M.~Dikaiakos.
\newblock Deliverable 3.1: Mandola monitoring dashboard.
\newblock Technical report, MANDOLA EU REC Project, September 2016.

\bibitem{paschalides2020mandola}
Demetris Paschalides, Dimosthenis Stephanidis, Andreas Andreou, Kalia Orphanou,
  George Pallis, Marios~D Dikaiakos, and Evangelos Markatos.
\newblock Mandola: A big-data processing and visualization platform for
  monitoring and detecting online hate speech.
\newblock {\em ACM Transactions on Internet Technology (TOIT)}, 20(2):1--21,
  2020.

\bibitem{qian2019benchmark}
Jing Qian, Anna Bethke, Yinyin Liu, Elizabeth Belding, and William~Yang Wang.
\newblock A benchmark dataset for learning to intervene in online hate speech.
\newblock {\em arXiv preprint arXiv:1909.04251}, 2019.

\bibitem{radford2019language}
Alec Radford, Jeffrey Wu, Rewon Child, David Luan, Dario Amodei, and Ilya
  Sutskever.
\newblock Language models are unsupervised multitask learners.
\newblock {\em OpenAI Blog}, 1(8), 2019.

\bibitem{richards2000counterspeech}
Robert~D Richards and Clay Calvert.
\newblock Counterspeech 2000: A new look at the old remedy for bad speech.
\newblock {\em BYU L. Rev.}, page 553, 2000.

\bibitem{riedl2020downsides}
Martin~J Riedl, Gina~M Masullo, and Kelsey~N Whipple.
\newblock The downsides of digital labor: Exploring the toll incivility takes
  on online comment moderators.
\newblock {\em Computers in Human Behavior}, 107:106262, 2020.

\bibitem{eisf2014}
Nanjira Sambuli and Kagonya Awori.
\newblock Monitoring online dangerous speech in kenya. insights from the umati
  project.
\newblock In Vazquez~Llorente R. and I.~Wall, editors, {\em Communications
  technology and humanitarian delivery: challenges and opportunities for
  security risk management}. European Interagency Security Forum, 2014.

\bibitem{DBLP:conf/evalita/SanguinettiCNFS20}
Manuela Sanguinetti, Gloria Comandini, Elisa~Di Nuovo, Simona Frenda, Marco
  Stranisci, Cristina Bosco, Tommaso Caselli, Viviana Patti, and Irene Russo.
\newblock Haspeede 2 @ {EVALITA2020:} overview of the {EVALITA} 2020 hate
  speech detection task.
\newblock In Valerio Basile, Danilo Croce, Maria~Di Maro, and Lucia~C. Passaro,
  editors, {\em Proceedings of the Seventh Evaluation Campaign of Natural
  Language Processing and Speech Tools for Italian. Final Workshop {(EVALITA}
  2020), Online event, December 17th, 2020}, volume 2765 of {\em {CEUR}
  Workshop Proceedings}. CEUR-WS.org, 2020.

\bibitem{schieb2016governing}
Carla Schieb and Mike Preuss.
\newblock Governing hate speech by means of counterspeech on facebook.
\newblock In {\em 66th ica annual conference, at Fukuoka, Japan}, pages 1--23,
  2016.

\bibitem{serban2016building}
Iulian~V Serban, Alessandro Sordoni, Yoshua Bengio, Aaron Courville, and Joelle
  Pineau.
\newblock Building end-to-end dialogue systems using generative hierarchical
  neural network models.
\newblock In {\em Thirtieth AAAI Conference on Artificial Intelligence}, 2016.

\bibitem{shang2015neural}
Lifeng Shang, Zhengdong Lu, and Hang Li.
\newblock Neural responding machine for short-text conversation.
\newblock In {\em Proceedings of the 53rd Annual Meeting of the Association for
  Computational Linguistics and the 7th International Joint Conference on
  Natural Language Processing (Volume 1: Long Papers)}, pages 1577--1586, 2015.

\bibitem{silverman2016impact}
Tanya Silverman, Christopher~J Stewart, Jonathan Birdwell, and Zahed Amanullah.
\newblock The impact of counter-narratives.
\newblock {\em Institute for Strategic Dialogue}, pages 1--54, 2016.

\bibitem{sordoni2015neural}
Alessandro Sordoni, Michel Galley, Michael Auli, Chris Brockett, Yangfeng Ji,
  Margaret Mitchell, Jian-Yun Nie, Jianfeng Gao, and Bill Dolan.
\newblock A neural network approach to context-sensitive generation of
  conversational responses.
\newblock In {\em Proceedings of the 2015 Conference of the North American
  Chapter of the Association for Computational Linguistics: Human Language
  Technologies}, pages 196--205, 2015.

\bibitem{Specia10AMTA}
Lucia Specia and Atefeh Farzindar.
\newblock Estimating machine translation post-editing effort with hter.
\newblock In {\em AMTA 2010- workshop, Bringing MT to the User: MT Research and
  the Translation Industry}. The Ninth Conference of the Association for
  Machine Translation in the Americas, The Ninth Conference of the Association
  for Machine Translation in the Americas, nov 2010.

\bibitem{stella2019influence}
Massimo Stella, Marco Cristoforetti, and Manlio De~Domenico.
\newblock Influence of augmented humans in online interactions during voting
  events.
\newblock {\em PloS one}, 14(5):e0214210, 2019.

\bibitem{DBLP:conf/konvens/StrussSRWK19}
Julia~Maria Stru{\ss}, Melanie Siegel, Josef Ruppenhofer, Michael Wiegand, and
  Manfred Klenner.
\newblock Overview of germeval task 2, 2019 shared task on the identification
  of offensive language.
\newblock In {\em Proceedings of the 15th Conference on Natural Language
  Processing, {KONVENS} 2019, Erlangen, Germany, October 9-11, 2019}, 2019.

\bibitem{tekiroglu2020generating}
Serra~Sinem Tekiro{\u{g}}lu, Yi-Ling Chung, and Marco Guerini.
\newblock Generating counter narratives against online hate speech: Data and
  strategies.
\newblock In {\em Proceedings of the 58th Annual Meeting of the Association for
  Computational Linguistics}, pages 1177--1190, Online, July 2020. Association
  for Computational Linguistics.

\bibitem{turchi-etal-2013-coping}
Marco Turchi, Matteo Negri, and Marcello Federico.
\newblock Coping with the subjectivity of human judgements in {MT} quality
  estimation.
\newblock In {\em Proceedings of the Eighth Workshop on Statistical Machine
  Translation}, pages 240--251, Sofia, Bulgaria, August 2013. Association for
  Computational Linguistics.

\bibitem{VinyalsL15}
Oriol Vinyals and Quoc~V. Le.
\newblock A neural conversational model.
\newblock {\em CoRR}, abs/1506.05869, 2015.

\bibitem{wang-2018-sketching}
Vincent Wang.
\newblock Sketching a {C}hinese writer{'}s vocabulary profile in {E}nglish: the
  case of {H}a {J}in.
\newblock In {\em Proceedings of the 32nd Pacific Asia Conference on Language,
  Information and Computation}, Hong Kong, 1{--}3 December 2018. Association
  for Computational Linguistics.

\bibitem{wright2017vectors}
Lucas Wright, Derek Ruths, Kelly~P Dillon, Haji~Mohammad Saleem, and Susan
  Benesch.
\newblock Vectors for counterspeech on twitter.
\newblock In {\em Proceedings of the First Workshop on Abusive Language
  Online}, pages 57--62, 2017.

\bibitem{DBLP:conf/semeval/ZampieriNRAKMDP20}
Marcos Zampieri, Preslav Nakov, Sara Rosenthal, Pepa Atanasova, Georgi
  Karadzhov, Hamdy Mubarak, Leon Derczynski, Zeses Pitenis, and {\c{C}}agri
  {\c{C}}{\"{o}}ltekin.
\newblock Semeval-2020 task 12: Multilingual offensive language identification
  in social media (offenseval 2020).
\newblock In Aur{\'{e}}lie Herbelot, Xiaodan Zhu, Alexis Palmer, Nathan
  Schneider, Jonathan May, and Ekaterina Shutova, editors, {\em Proceedings of
  the Fourteenth Workshop on Semantic Evaluation, SemEval@COLING 2020,
  Barcelona (online), December 12-13, 2020}, pages 1425--1447. International
  Committee for Computational Linguistics, 2020.

\bibitem{10.1145/3209978.3210080}
Yaoming Zhu, Sidi Lu, Lei Zheng, Jiaxian Guo, Weinan Zhang, Jun Wang, and Yong
  Yu.
\newblock Texygen: A benchmarking platform for text generation models.
\newblock In {\em The 41st International ACM SIGIR Conference on Research and
  Development in Information Retrieval}, SIGIR '18, page 1097–1100, New York,
  NY, USA, 2018. Association for Computing Machinery.

\end{thebibliography}

\clearpage

\appendix

\section{Usability Task Instructions}
\label{sec:usability_tasks_instruction}

\begin{table}[h]
\caption{Instructions for Usability Tasks}
\label{tab:usability_tasks}
\centering
\begin{tabularx}{\textwidth}{l|X}
\hline
\textbf{Task} & \textbf{Instruction} \\ \hline
Hate  distribution & Please choose two hashtags and a temporal framework of your own choice. Now analyse the network of hate speakers. What is their shape? What is the distribution of hate speech? Are there main nodes (i.e. influencers) which lead the flow of hate speech? Is the net wide or narrow? In your opinion, what network is the most dangerous in terms of anti-Muslim hate speech and why? \\ \hline
Hate  timeline & In the ``Hashtag trends'' section, please first select one hashtag of your own choice and, subsequently, a temporal window from the drop-down menu (snapshot). Then maintain the same hashtag, changing the snapshot. How do trends vary across time with regard to that specific hashtag? Does the shape of the network change? \\ \hline
Islamisation analysis & Starting from the `Hashtag trends' window, please analyse the key hashtags and keywords related to the concept of `Islamisation', checking the data obtained from the network analysis and the graphs that the platform accomplishes. Later, please identify the hate speakers who mostly utilize the hashtags and keywords that you have identified. \\ \hline
Recent trends & Please select one of the most recent Islamophobic hashtags and analyse the context in which it is used: does it refer to the national or the international context? What political actors are involved? Are the hate speakers who are active closely affiliated to any particular political party?  \\ \hline
\end{tabularx}
\end{table}

\end{document}